\documentclass[conference]{IEEEtran}
\IEEEoverridecommandlockouts

\usepackage{cite}
\usepackage{amsmath,amssymb,amsfonts}
\usepackage{algorithmic}
\usepackage{graphicx}
\usepackage{textcomp}
\usepackage{xcolor}
\usepackage{multicol}

\def\BibTeX{{\rm B\kern-.05em{\sc i\kern-.025em b}\kern-.08em
    T\kern-.1667em\lower.7ex\hbox{E}\kern-.125emX}}
\begin{document}

\fontsize{9}{11}\selectfont

\title{Preference Alignment Improves Language Model-Based TTS
\thanks{
  Jinchuan and Jiatong are interns at Tencent AI LAB during this work. Code is released: https://github.com/espnet/espnet/blob/speechlm/egs2/librispeech/ speechlm1/local/train\_hfrl.sh
  }
}

\author{\textit{Jinchuan Tian$^{1}$,
Chunlei Zhang$^{2}$,
Jiatong Shi$^{1}$,
Hao Zhang$^2$,
Jianwei Yu$^2$,
Shinji Watanabe$^1$,
Dong Yu$^2$} \\
\\
$^{1}$Language Technologies Institute, Carnegie Mellon University, $^{2}$Tencent AI LAB \\
\footnotesize\texttt{\{jinchuat, jiatongs, swatanab\}@andrew.cmu.edu, \{cleizhang, aaronhzhang, tomasyu, dyu\}@tencent.com} \\
}

\maketitle

\begin{abstract}
Recent advancements in text-to-speech (TTS) have shown that language model (LM)-based systems offer competitive performance to their counterparts. Further optimization can be achieved through preference alignment algorithms, which adjust LMs to align with the preferences of reward models, enhancing the desirability of the generated content. This study presents a thorough empirical evaluation of how preference alignment algorithms, particularly Direct Preference Optimization (DPO), enhance LM-based TTS. With a 1.15B parameter LM-based TTS model, we demonstrate that preference alignment consistently improves intelligibility, speaker similarity, and proxy subjective evaluation scores, with the latter two metrics surpassing even human speech in certain evaluations. We also show preference alignment is applicable to low-resource scenarios and effectively generalized to out-of-domain applications. 
\end{abstract}


\section{Introduction}
Text-to-speech (TTS) aims to synthesize human speech from the given text and, optionally, non-text conditions \cite{survey}. Traditionally, mainstream TTS systems operate in continuous space \cite{fastspeech, tacotron, vits}. 
Recent advancements in audio coding have enabled high-quality audio tokenization \cite{soundstream, encodec, du2024funcodec, zhang2024speechtokenizer, espnet_codec}. The tokenization allows TTS models to function effectively in discrete space \cite{discrete_challenge}, particularly through the use of language model (LM)-based approaches \cite{valle, speartts, yang2024uniaudio, seedtts, vallex, audiolm, audiopalm, soundstorm, voxtlm}. 
LM-based approaches are featured for the simplified training and inference pipeline, enabling the model to learn the relationships between input and output sequences more efficiently.
These systems have gained popularity, achieving state-of-the-art performance by scaling up both data volume and parameter sizes \cite{audiolm_survey}. They also exhibit remarkable zero-shot capabilities in tasks such as speaker identity cloning \cite{valle} and cross-lingual synthesis \cite{vallex}.

Despite these advances, generating high-quality, natural-sounding speech requires not only scaling up but also being aligned with human perception.
Preference alignment (PA) is a set of training algorithms widely employed in text-based LM development. The goal of PA is to align model outputs with specific preferences, which are abstract and challenging to learn by maximize-a-posterior (e.g., cross-entropy loss) \cite{hfrl_survey}. 
Typically formulated as a reinforcement learning problem, PA first models the preferences by a reward model and then uses the reward model to guide LMs toward generating content that maximizes reward values \cite{ppo, instructgpt}. When these preferences are derived from humans, the process is widely known as human feedback reinforcement learning (HFRL). Recent advancements in PA allow for solving the optimization problem in a closed form, eliminating the need for explicit reward modeling, which significantly simplifies and stabilizes training \cite{dpo, kto, simpo, ipo, orpo}. PA (or HFRL) is verified effective in understanding and following highly abstract preference (e.g., human value), and has become a common practice to ensure that text LMs exhibit desirable traits, such as helpfulness, truthfulness, and harmlessness \cite{gpt4, llama3.1}.

Although preference alignment methods are widely adopted in text LLM development, they are less explored in the speech/audio community, particularly the LM-based TTS. SpeechAlign \cite{speechalign} explored multiple preference alignment methods on LM-based TTS, but only used ground truth as the positive examples. UNO \cite{uno} optimized on unpaired preference data and considers the annotation uncertainty in subjective evaluation. RIO \cite{rio} leverages the Bayesian principle to select preference data. 
It is reported that industrial systems, such as SeedTTS \cite{seedtts}, adopt DPO and PPO in their human preference alignment stage.
Besides TTS, TANGO2 \cite{majumder2024tango} applies DPO to diffusion-based text-to-audio systems.

In this work, we apply a PA objective, direct performance optimization (DPO) \cite{dpo}, to LM-based TTS systems, guiding them to generate speech that is preferred across multiple metrics. 
Although prior works have preliminarily verified the feasibility of PA in TTS, this work provides transparency and details details in its implementation. 
We address multiple key practical issues, including 
(1) preference pair selection; 
(2) hyper-parameter search; 
(3) effect of length normalization; 
(4) metric selection; 
(5) effect of supervised fine-tuning (SFT); 
(6) label efficiency; 
(7) iterative optimization; 
(8) out-of-domain evaluation. Our baseline model, with 1.15B parameters, is trained on 55k hours of open-source English speech data. We demonstrate that applying PA to this baseline model significantly improves its intelligibility, speaker similarity, and proxy subjective evaluation scores, even outperforming human ground truth in the latter two metrics. 
Additionally, we show that preference alignment can be implemented with as little as 1 hour of data, and its improvement can be effectively generalized to out-of-domain scenarios.

\section{Preference Alignment on Language Model-Based TTS}
\subsection{Language Model-Based TTS}
TTS is a conditional generation task that generates speech signal $\mathbf{y}$ based on the given conditions $\mathbf{x}$, such as input text string $\mathbf{s}$ and other non-textual cues. For simplicity, this work assumes a short clip from the same speaker of $\mathbf{y}$, i.e., $\mathbf{y}_{\text{ref}}$, is the only non-textual cue, from which features like speaker identity and prosody can be imitated. Thus, the training objective of TTS is to maximize the posterior:
\begin{flalign}
   \textstyle \max_\theta P_{\theta}(\mathbf{y}|\mathbf{x}) = \max_\theta P_{\theta}(\mathbf{y}|\mathbf{s}, \mathbf{y}_{\text{ref}})
\end{flalign}
where $\theta$ is the trainable parameter of the model.

In the context of discrete space modeling, particularly LM-based TTS, all audio $\mathbf{y}$, $\mathbf{y}_{\text{ref}}$ can be converted into discrete codes by audio codec encoding \cite{soundstream, encodec, du2024funcodec, zhang2024speechtokenizer, espnet_codec}, s.t., $\mathbf{y}^{\text{d}}, \mathbf{y}^{\text{d}}_{\text{ref}}$. Text input $\mathbf{s}$ can also be tokenized into a integer vector $\mathbf{s}^{\text{d}}$. By splicing $\mathbf{s}^{\text{d}}$ with $\mathbf{y}^{\text{d}}_{\text{ref}}$ and $\mathbf{y}^{\text{d}}$, the example sequence $[\mathbf{s}^{\text{d}}, \mathbf{y}^{\text{d}}_{\text{ref}}, \mathbf{y}^{\text{d}}]$ is formed and then learned by a language model. Cross-entropy loss is applied to $\mathbf{y}^{\text{d}}$ so that the posterior $P_\theta(\mathbf{y}^{\text{d}}|\mathbf{s}^{\text{d}}, \mathbf{y}^{\text{d}}_{\text{ref}})$ is maximized. During inference, the predicted sequence $\hat{\mathbf{y}}^{\text{d}}$ is first generated by an LM, and then the output speech $\hat{\mathbf{y}}$ can be reconstructed from it using audio codec decoding. 


Usually, audio codec models tokenized each frame of audio into $n_q$ codes ($n_q>1$), which makes the example sequence $[\mathbf{s}^{\text{d}}, \mathbf{y}^{\text{d}}_{\text{ref}}, \mathbf{y}^{\text{d}}] \in \mathbb{Z}^{T\times n_q}$ a two-dimensional sequence\footnote{$\mathbf{s}^{\text{d}}$ is repeated or padded to two-dimensional.}. $T$ stands for number of frames. As standard LMs work with one-dimensional sequence, modeling the sequences with the extra $n_q$-dimension is non-trivial \cite{musicgen}. This work adopt Multi-Scale Transformer \cite{yang2024uniaudio} as the model architecture, which first uses a global Transformer to predict an embedding for each audio frame; and then a local Transformer predicts the $n_q$ codes sequentially based on each frame embedding. Both global and local Transformers are causal. Like standard LM, Multi-Scale Transformer also predicts the code-level posterior for each audio code within each frame, which is then used in loss computing (e.g., cross-entropy) and model inference.
For simplicity, in the rest of this paper, we re-name the conditional sequence as $\textbf{x}=[\textbf{s}^{\text{d}}, \textbf{y}^{\text{d}}_{\text{ref}}]$ and the target sequence $\mathbf{y}=[\mathbf{y}^\text{d}]$, both are in discrete space.

\subsection{Preference Alignment}
Cross-entropy objective in LM-based TTS training is to maximize the posterior of target sequence $\mathbf{y}$. However, higher posterior in $\mathbf{y}$ (and the corresponding waveform reconstructed from it) does not necessarily lead to content that is more preferred by human perception or other proxy metrics \cite{human_like}. Alternatively, PA is an approach that directly optimizes the LM toward these preferences and thus improves the sample quality \cite{hfrl_survey}.

\noindent \textbf{Problem Formulation:} PA is usually described as a reinforcement learning problem: assume there is a latent reward model $r^*(\mathbf{x}, \mathbf{y})$ that represents the preferences by a scalar reward, higher means more preferred. Thus, with the given reward model, the optimization objective is to guide the LM to pursue a higher expected reward:
\begin{flalign}
\max_{\theta}\mathbb{E}_{\mathbf{y}\sim P_\theta(\mathbf{y}|\mathbf{x})}[r(\mathbf{x}, \mathbf{y})] - \beta\cdot\mathbb{D}_{KL}[P_\theta(\mathbf{y}|\mathbf{x}) || P_{\text{ref}}(\mathbf{y}|\mathbf{x})] \label{init_def}
\end{flalign}
where the latter term is a KL-divergence constraint to avoid the LM $P_{\theta}$ drifting too far away from a reference model $P_{\text{ref}}$. $\beta$ is a hyper-parameter, larger means stronger constraint. The choice of $\beta$ is explored in Sec.\ref{beta_selection}. In common practice, the reference model $P_{\text{ref}}$ is initialized identically with $P_\theta$ and is frozen during training.

Conventionally, the optimization in Eq. \eqref{init_def} works with an explicit reward model \cite{instructgpt}. As the latent reward model is usually unavailable, a proxy reward model $r_\phi(\mathbf{x}, \mathbf{y})$ is first built from the preference dataset instead. Subsequently, the optimization is conducted using proximal policy optimization (PPO) \cite{ppo}. This workflow is complicated and PPO sometimes encounters instability issues in training~\cite{zhu2023fine}. Recent advances in PA demonstrate that, under certain circumstances, the optimization in Eq.~\eqref{init_def} can be solved in close form without building an explicit reward model. A representative approach is Direct Preference Optimization (DPO) \cite{dpo}.

\noindent  \textbf{Direct Preference Optimization (DPO):}
DPO deals with a special case where the preference data is win-lose pairs: with the same conditions $\mathbf{x}$, the probability of $\mathbf{y}_{\text{w}}$ being more preferred than sequence $\mathbf{y}_{\text{l}}$ follows Bradley-Terry model \cite{btmodel}:
\begin{flalign}
P(\mathbf{y}_{\text{w}} > \mathbf{y}_{\text{l}} |\mathbf{x}) = \frac{\text{exp}(r^*(\mathbf{x}, \mathbf{y}_{\text{w}}))
}{\text{exp}(r^*(\mathbf{x}, \mathbf{y}_{\text{w}})) + \text{exp}(r^*(\mathbf{x}, \mathbf{y}_{\text{l}}))}
\end{flalign}

So that, with the known preference data triplets ($\mathbf{x}, \mathbf{y}_{\text{w}}$, $\mathbf{y}_{\text{l}}$), an explicit proxy reward model $r_\phi$ can be trained by maximum-a-likelihood criterion, with $\sigma$ being the \textit{sigmoid} function:
\begin{flalign}
    \max_{r_\phi}\mathbb{E}[\log \sigma(r_\phi(\mathbf{x}, \mathbf{y}_{\text{w}}) - r_\phi(\mathbf{x}, \mathbf{y}_{\text{l}}))]
    \label{train_reward}
\end{flalign}

Also, it has been proved that, in Eq.~\eqref{init_def}, the LM $P_\theta(\mathbf{y}|\mathbf{x})$ becomes optimal if and only if:
\begin{flalign}
    r_\phi(\mathbf{x}, \mathbf{y}_{\text{w}}, \mathbf{y}_{\text{l}}) = \beta \cdot \frac{P_\theta(\mathbf{y}|\mathbf{x})}{P_{\text{ref}}(\mathbf{y}|\mathbf{x})} + \beta \cdot \mathbb{Z}(\mathbf{x})
    \label{reward_solution}
\end{flalign}
where $\mathbb{Z}(\mathbf{x})$ is termed as partition function that is independent of the generation target $\mathbf{y}$. 

Finally, substituting Eq.~\eqref{reward_solution} into Eq.~\eqref{train_reward} excludes the reward model $r_\phi(\mathbf{x}, \mathbf{y})$; training the explicit reward models is then transformed as direct optimization over the LM $P_\theta(\mathbf{y}|\mathbf{x})$:
\begin{flalign}
    \mathit{L}_{\text{DPO}} = -\mathbb{E}\left[\log \sigma\left(\beta \cdot \log \frac{P_\theta(\mathbf{y}_{\text{w}}|\mathbf{x})}{P_{\text{ref}}(\mathbf{y}_{\text{w}}|\mathbf{x})} - \beta \cdot \log \frac{P_\theta(\mathbf{y}_{\text{l}}|\mathbf{x})}{P_{\text{ref}}(\mathbf{y}_{\text{l}}|\mathbf{x})}\right)\right]
    \label{dpo}
\end{flalign}

\noindent\textbf{DPO on LM-based TTS:} 
Our DPO training starts from a baseline LM-based TTS model pre-trained with cross-entropy loss (detailed in Sec. \ref{experimental_setup}). Specifically, any posterior $P(\mathbf{y}|\mathbf{x})$ in Eq.~\eqref{dpo} are computed by flattening the two-dimensional $\textbf{y}$ into row-first sequence and then summing the code-level log-posterior in auto-regressive format. To align with human perception, it would be ideal if the preference data pairs $(\mathbf{x}, \mathbf{y}_\text{w}, \mathbf{y}_\text{l})$ can come from human labelers. Instead, this work adopts several pre-trained metric models as the proxy of real human preferences. With the same condition ${\mathbf{x}}$, utterances are first scored by these models; the utterances with better/worse scores are set to $\mathbf{y}_{\text{w}}$ and $\mathbf{y}_{\text{l}}$, respectively. These metric models are also
detailed in Sec. \ref{experimental_setup}. $\mathbf{y}_{\text{w}}$ and $\mathbf{y}_{\text{l}}$ can be either generated from the LM or from natural speech, which is explored in Sec.\ref{pair_selection}.


\section{Experiments}
\subsection{Experimental Setup}
\label{experimental_setup}
\noindent\textbf{Data, Task Setup, and Tokenization: } 
We build our baseline model with LibriSpeech \cite{librispeech-corpus}, GigaSpeech \cite{gigaspeech} and the English part of Multilingual LibriSpeech \cite{pratap2020mls}, summing up to around 55k hours. 
Following \cite{valle}, speaker IDs are always available for all datasets and are used to select a 3-second speech clip from the same speaker\footnote{Speaker IDs of GigaSpeech are generated by AutoPrep \cite{autoprep}.}. All input text is tokenized into phone sequences by \texttt{g2p-en}\footnote{https://github.com/Kyubyong/g2p} before language modeling. We adopt our reproduced SoundStream \cite{soundstream} model for audio tokenization, which is configured as 50 frames per second and 8 codes per frame, i.e., $n_q = 8$.

\noindent\textbf{Model:} 
We adopt Multi-Scale Transformer \cite{yang2024uniaudio} as the model architecture.
The global Transformer has 25 layers, each of which has an attention size of 1600, a feedforward size of 6400, and 25 attention heads. Those numbers for the local Transformer are \{6, 384, 1536, 6\} respectively. The total trainable parameters are 1.15B.

\noindent\textbf{Training and Inference:} The baseline model is updated by 1M steps with the global batch size of around 80k frames. AdamW optimizer \cite{adamw} with a peak learning rate of 2e-4 is adopted, with 70k warmup steps, and then decays exponentially. Training is based on 8$\times$A100-40G GPUs. 
For inference, we settled down with top-$k$ sampling using $k=30$; we re-scale the logits with a temperature of 1.2. For each example, we perform batch inference with the size of 10 using the same condition (text and reference speech clips). We do not introduce any human prior in the selection of reference speech clips as they are usually provided by users.

\noindent\textbf{Metric Models and Evaluation: }
For the LM-based TTS system, we are interested in three metrics of the generated content: intelligibility (WER), speaker similarity (SPK\_SIM), and proxy subjective evaluation scores (Proxy MOS). 
The specific models for each metric are: Whisper-large \cite{whisper} for WER; Speaker embeddings from RawNet \cite{jung19b_interspeech, jung24c_interspeech} for SPK\_SIM; UTMOS \cite{utmos} for Proxy MOS. These metric models are also used in most evaluations. We use additional metric models to ensure the TTS model is improved in general rather than over-fits to the preference of these pre-trained metric models (sec. \ref{out_of_domain}). 
We adopt LibriSpeech Test-Clean in most evaluations while VCTK \cite{vctk} is for out-of-domain scenarios.
Although re-ranking among the batch-generated examples can significantly improve the performance \cite{ralle}, this work does not include that operation and reports every number as the average of all 10 examples.

\subsection{Experiments and Analysis}
We first demonstrate the performance of our optimal model \texttt{E1} and our baseline model in table \ref{tab:baseline}. Although the baseline model already achieves comparable performance with popular systems, applying DPO still achieves significant improvement in all three metrics.  We detailed our exploration step-by-step as follows. 

\begin{table}[t]
\centering
    \caption{Performance overview of our baseline and DPO-trained system. All results are averaged from 10 generated utterances of each sample. The results of reference systems may not be comparable due to different evaluation protocols.}
    \begin{tabular}{l|c|c|c}
    \hline
         System & WER & SPK\_SIM & Proxy MOS  \\
         \hline
         Ground Truth & 1.8 & 0.625 & 4.08 \\
         \hline
         Baseline (ours) & 4.5 & 0.635 & 3.80 \\
         Baseline + DPO (\texttt{E1}, ours) & \textbf{3.0} & \textbf{0.667} & 4.23 \\
         \hline
         Reference - ChatTTS  & 8.3 & - & 3.46 \\
         Reference - YourTTS\cite{yourtts} & 7.7 & 0.337 & 3.45 \\
         Reference - Vall-E\cite{valle}   & 5.9 & 0.580 & \textbf{4.38} \\
         \hline
    \end{tabular}
    \vspace{-10pt}
    \label{tab:baseline}
\end{table}

All PA experiments are based on LibriSpeech. As LibriSpeech is already included in baseline model training, this setup excludes the impact of introducing unseen high-quality data. We conduct inference on the whole train-960 set for follow-up preference data curation.
Our exploration is based on DPO with a conservative setup in the initial trials: a constant learning rate of 3e-7 and $\beta=0.1$. Empirically, we find DPO is sensitive to the number of updates, so we use a large batch size to ensure only 350 updates are made within one epoch.

\subsubsection{Preference Pair Selection}
\label{pair_selection}
We examine two solutions to how preference data pairs are curated. \texttt{A2}: use ground truth as $\mathbf{y}_w$ and a randomly selected generated example as $\mathbf{y}_l$. This assumes the ground truth always outperforms generated examples. \texttt{B2}: rank all generated samples using certain preference metrics; select the top 20\% best and worst examples as $\mathbf{y}_w$ and $\mathbf{y}_l$ respectively. For simplicity, we only use SPK\_SIM to rank these examples.

\begin{figure}[h]
    \centering
    \vspace{-5pt}
    \includegraphics[width=0.5\textwidth]{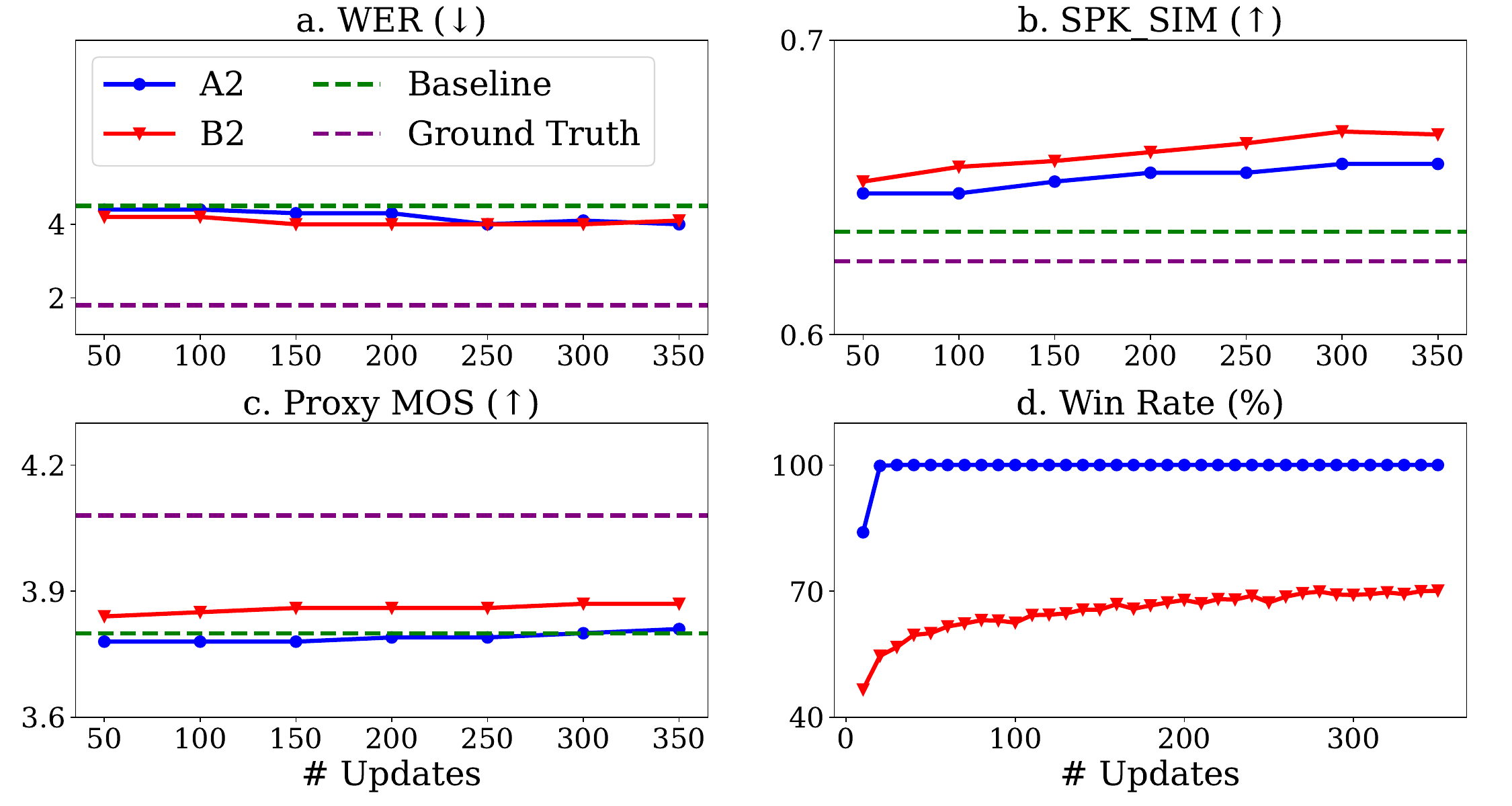}
    \caption{Comparison of different data curation strategies}
    \vspace{-10pt}
    \label{fig:data_curation}
\end{figure}

We evaluate \texttt{A2} and \texttt{B2} by every 50 updates, the results are in Fig.\ref{fig:data_curation}. It is clear that \texttt{B2} outperforms the baseline and \texttt{A2}, especially on the SPK\_SIM and Proxy MOS metrics. We further compare the win rate\footnote{
  Win rate is the ratio that $r_\phi(\mathbf{x}, \mathbf{y}_{w}) > r_\phi(\mathbf{x}, \mathbf{y}_{l})$, which is usually used to monitor the progress of the optimization problem in Eq.~\eqref{train_reward}.
}
of \texttt{A2} and \texttt{B2}. As suggested in Fig.\ref{fig:data_curation}.d, the win rate of \texttt{A2} reached 99.8\% only after 20 updates, which indicates that there is a trivial difference between the natural speech and generated speech in discrete space, making the model less explore the features that can improve the model performance. By contrast, since both $\mathbf{y}_w$ and $\mathbf{y}_l$ are generated in \texttt{B2}, the optimization is non-trivial and provides better performance.

\subsubsection{Hyper-Parameter Search}
\label{beta_selection}
\begin{figure}[h]
    \centering
    \vspace{-8pt}
    \includegraphics[width=0.5\textwidth]{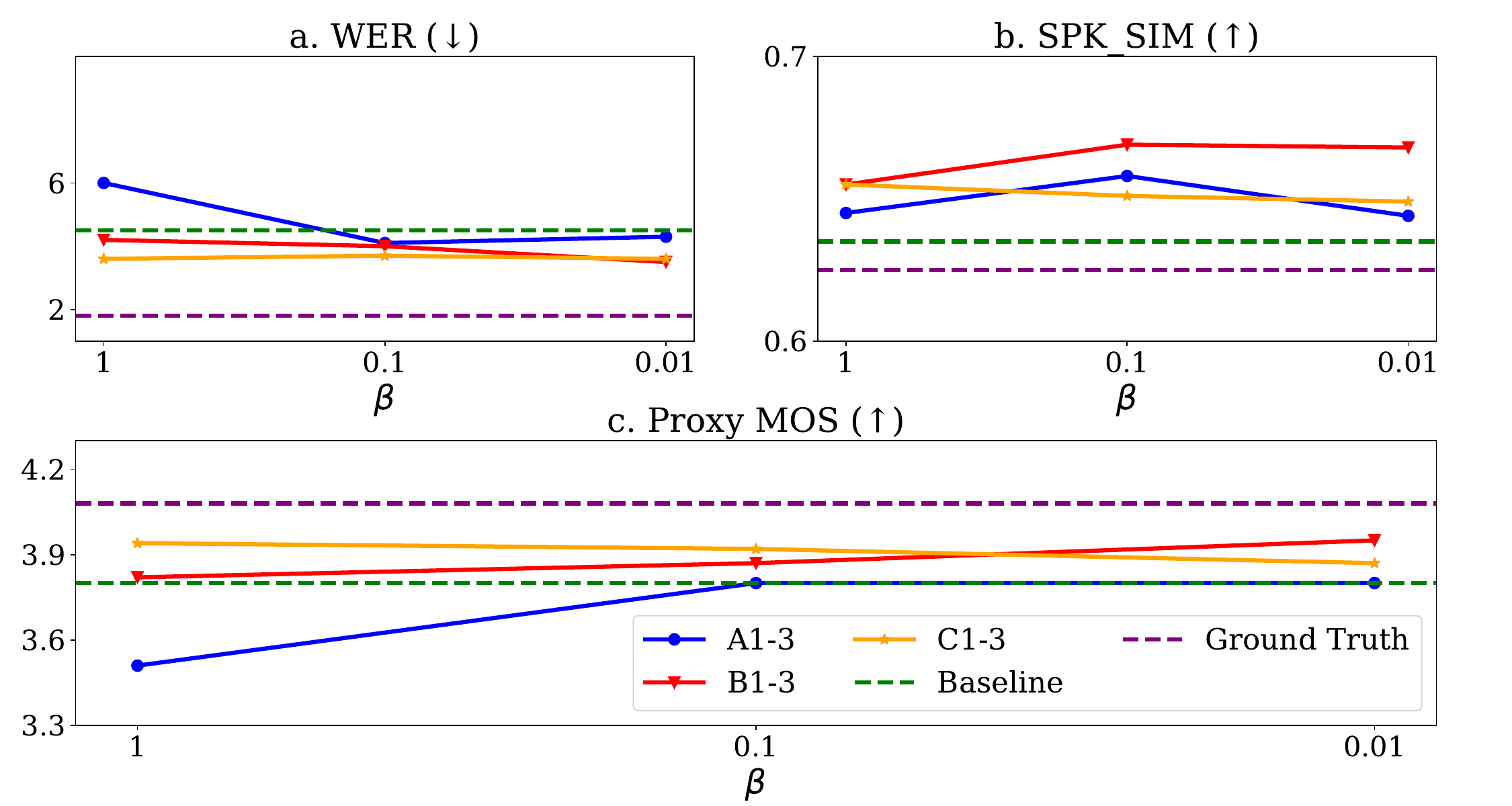}
    \caption{Comparison of different $\beta$ choices and w/o length normalization.}
    \label{fig:beta}
\end{figure}

Based on \texttt{A2} and \texttt{B2}, we extend to \texttt{A1} and \texttt{B1} with $\beta=1$; \texttt{A3} and \texttt{B3} with $\beta=0.01$. Fig.\ref{fig:data_curation} suggests that the results achieved by 300 updates are nearly optimal, so we only evaluate the models with the same number of updates. The results are in Fig.\ref{fig:beta}.

As \texttt{A1-3} consistently under-performs \texttt{B1-3}, we proceed with both $\mathbf{y}_w$ and $\mathbf{y}_l$ being generated by baseline. \texttt{B3} outperforms \texttt{B1} and \texttt{B2} consistently, so we proceed with $\beta=0.01$.

\subsubsection{Effect of Length Normalization}
It is mentioned in \cite{seedtts} that standard DPO will lengthen the generated examples while \cite{simpo} mentioned that length normalization can be used for regularization. So we applied length normalization to all posteriors in Eq.~\eqref{dpo} and did experiments \texttt{C1-3} with $\beta=\{1, 0.1, 0.01\}$. The results are in Fig.\ref{fig:beta}. 

We find that DPO with length normalization can consistently improve the baseline model. Since \texttt{C1-3} under-perform \texttt{B3}, we still proceed without length normalization. Additionally, we observe that the examples generated by \texttt{B3} are 5.1\% longer than the ground truth, while that number for \texttt{C3} is 4.1\%, so that, applying length normalization does not have a noticeable impact on the lengths of the generated examples. We also observe in \texttt{C1-3} that applying length normalization increases the robustness toward $\beta$ choices.

\subsubsection{Metric Selection}
So far we only selected the preference pairs by SPK\_SIM metric. We then change the metric to Proxy MOS (\texttt{D1}) and WER (\texttt{D2}). Additionally, we combine the ranking results of SPK\_SIM, Proxy MOS, and WER in a naive way\footnote{
With each metric, we rank all examples and assign scores from 0 to 9, lower is better. Examples with lower overall scores are preferred.
} (\texttt{D3}). The results are in Tab.\ref{tab:metric}.

\begin{table}[t]
    \centering
    \caption{Comparison of different data curation metrics and w/o SFT.}
    \scalebox{0.95}{
    \begin{tabular}{l|l|c|c|c|c}
    \hline
      Exp. & Metric & SFT &  WER & SPK\_SIM & Proxy MOS  \\
      \hline
      Ground Truth & - & - & 1.8 & 0.625 & 4.08 \\
      \hline
         Baseline  & - & - & 4.5 & 0.635 & 3.80 \\
      \texttt{B3}  & SPK\_SIM & & 3.5 & {0.668} & 3.95 \\
      \texttt{D1}  & WER  & & 3.6 & 0.653 & 3.95\\
      \texttt{D2}  & Proxy MOS & & 3.3 & 0.649 & {4.25} \\
      \texttt{D3}  & ALL  & & {3.1} & 0.663 & 4.20 \\
      \texttt{E1}  & ALL & \checkmark & \textbf{3.0} & \textbf{0.667} & \textbf{4.23} \\
      \hline
    \end{tabular}}
    \vspace{-10pt}
    \label{tab:metric}
\end{table}

As suggested in the table, adopting any metric for preference pair selection and then applying DPO can improve the baseline model on all three metrics consistently.
For Proxy MOS and SPK\_SIM, the optimal performance is achieved when the corresponding model is adopted for preference pair selection (\texttt{B3} and \texttt{D2}). Applying WER alone yields the worst WER result among all 4 DPO experiments (\texttt{D1}). We conjecture that the concept of WER focuses on the local errors within speech while DPO considers the whole sequences, which makes WER less ideal for DPO training. Using all metrics (\texttt{D3}) achieves encouraging and balanced performance on all metrics, so later on we proceed with \texttt{D3} setup.

\subsubsection{Effect of Supervised Fine-Tuning}
It is a common practice for the pre-trained model to experience supervised fine-tuning (SFT) before the preference alignment stage \cite{gpt4}. Thus, we fine-tune the baseline model on the $\textbf{y}_\text{w}$ of $\texttt{D3}$ for one epoch before the DPO training, using the same learning schedule as in Sec.\ref{experimental_setup} (\texttt{E1}). The results are in Tab.\ref{tab:metric}. It indicates that applying this SFT training to the baseline model provides marginal improvement after DPO training. For simplicity, we proceed without this SFT stage.

\subsubsection{Label Efficiency}
Our DPO training leverages the full LibriSpeech training set, which is overly abundant. We then reduce the training data volume to \{100, 10, 1\} hours. We conduct experiment with two setups. \texttt{F1-3}: reduce the batch size to 10\% of the original but keep the number of updates unchanged; \texttt{F4-6}: keep both batch size and number of updates unchanged. This change means the data will be used for more than one epoch in \texttt{F2-6}.

\begin{figure}[h]
    \centering
    \vspace{-8pt}
    \includegraphics[width=0.5\textwidth]{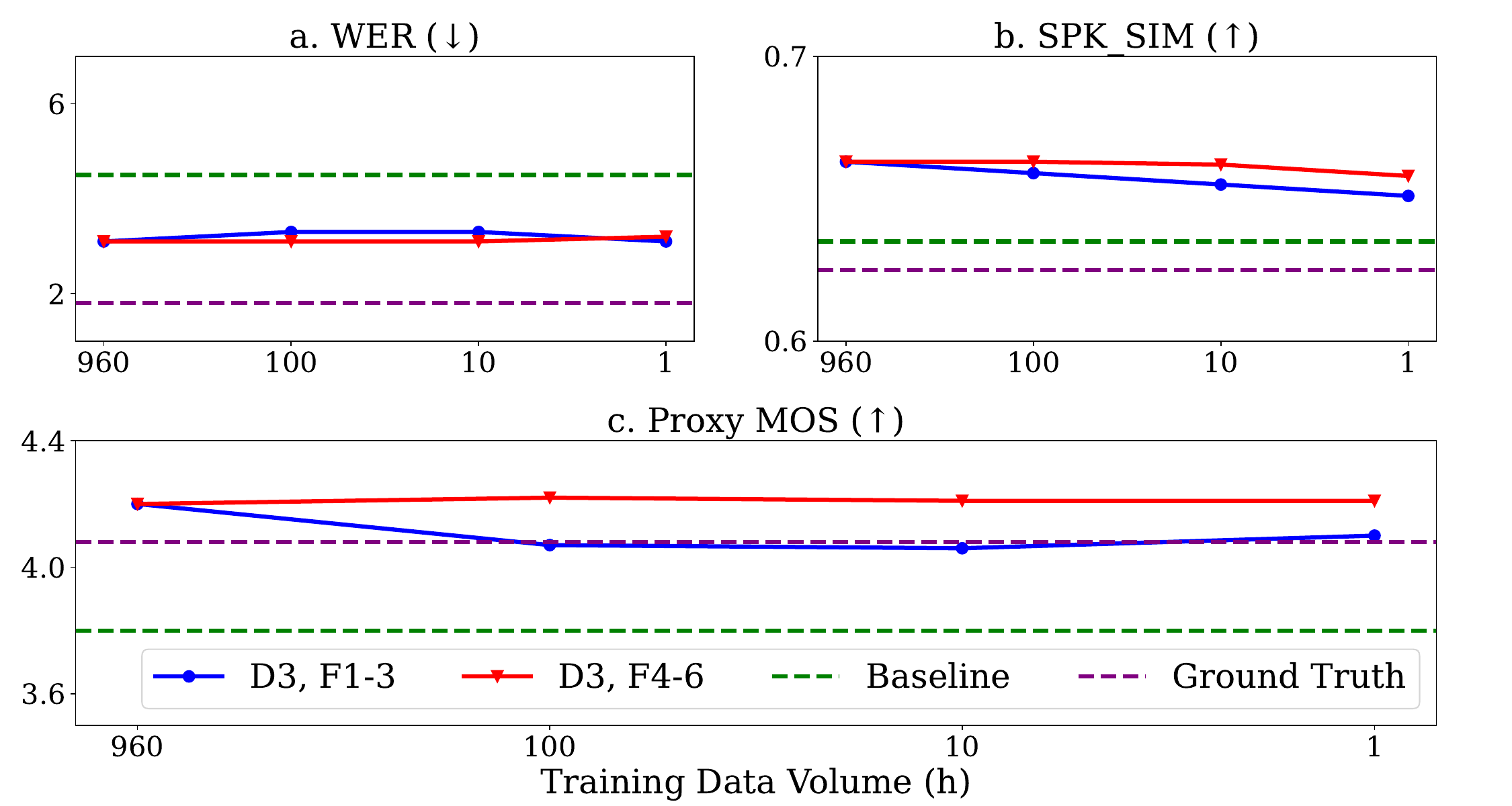}
    \caption{Comparison of different training data volume.}
    \label{fig:data_volume}
    \vspace{-8pt}
\end{figure}

We summarize the results in Fig.\ref{fig:data_volume}. The results of \texttt{F1-6} are all close to that of \texttt{D3}, which shows that DPO can work with preference pairs as small as 1 hour (258 examples specifically). Comparing \texttt{F1-3} to \texttt{F4-6}, using large batch size is slightly helpful in terms of SPK\_SIM and Proxy MOS.

\subsubsection{Iterative Optimization}
So far all experiments leverage the preference pairs generated by the baseline model. It would be more desirable if these pairs could be generated online by the model under training. As an approximation, iterative optimization \cite{llama3.1, iterative} is to repetitively generate preference pairs by the DPO-trained model in the last round and then leverage these pairs to train the next model. For cost reason, we examine iterative optimization only with the train-clean-100 subset. Starting from \texttt{F4}, we iterate the model for one more rounds, which yields \texttt{G1}. The results are shown in Table.\ref{tab:iter_unseen}.a. 
After multiple trails, we find the iterative optimization is fragile and we cannot achieve further improvement on \texttt{G1}.

\subsubsection{Out-of-Domain performance}
\label{out_of_domain} 

Given the success on the in-domain test set (LibriSpeech test-clean), we further show that our DPO training also improves out-of-domain performance. We evaluate \texttt{E1} on a subset of VCTK, which is not included in either baseline training or DPO training. As suggested in Tab.\ref{tab:iter_unseen}.b, DPO training achieves consistent improvement on all three metrics.

In Tab.\ref{tab:iter_unseen}.c, we evaluate the baseline and \texttt{E1} with three unseen metric models. The results suggest that the model is improved by DPO training in a general sense, rather than over-fitting on the metric models used in preference pair selection.

\begin{table}[h]
    \centering
    \vspace{-5pt}
    \caption{Evaluation on iterative optimization, unseen data domain, and unseen metric models}
    \scalebox{0.93}{
    \begin{tabular}{l|c|c|c}
    \hline
    \multicolumn{4}{l}{\textbf{a. Evaluation on iterative optimization}} \\ 
    \hline
    Exp.     & WER & SPK\_SIM & Proxy MOS \\
    \hline
    Baseline & 4.5 & 0.635 & 3.80 \\
    \texttt{F4} & \textbf{3.1} & \textbf{0.663} & \textbf{4.22} \\ 
    \texttt{G1} & 5.1 & 0.631 & 3.95 \\
    \hline\hline
    \multicolumn{4}{l}{\textbf{b. Evaluation with VCTK subset}} \\ 
    \hline
        Exp.     & WER & SPK\_SIM & Proxy MOS \\
       \hline
       Baseline & 1.6 & 0.677 & 3.87 \\
       \texttt{E1} & \textbf{1.5} & \textbf{0.688} & \textbf{4.15} \\
    \hline\hline
    \multicolumn{4}{l}{\textbf{c. Evaluation with unseen metric models}} \\ 
    \hline
             & WER & SPK\_SIM & Proxy MOS \\
       Exp.  & (OWSM v3.2 \cite{tian24_interspeech}) & (ECAPA-TDNN \cite{ecapa-tdnn}) & (DNSMOS \cite{dnsmos}) \\
       \hline
       Baseline & 5.0 &  0.655 & 3.90 \\
       \texttt{E1} & \textbf{3.2} & \textbf{0.679} & \textbf{4.00} \\
       \hline 
    \end{tabular}}
    \label{tab:iter_unseen}
\end{table}


\textbf{Summary:} Although there are many factors that can affect the performance of preference alignment, these algorithms are robust to configurations and can improve the LM-based TTS systems in most cases. Specifically, we find that using generated win-lose pairs and a small $\beta$ (such as 0.01) yields optimal performance. The benefits of using length normalization and SFT are marginal. Using all metrics to select preference pairs achieves balanced improvement in all directions; applying preference alignment iteratively is fragile. Preference alignment methods can work with as little as 1 hour of data; larger batch size provides a slight extra improvement. Finally, our \texttt{ E1} model with DPO outperforms the ground truth human speech in both SPK\_SIM and Proxy MOS metrics.

\section{Conclusion}
Considering the prosperity of LM-based approaches in TTS research, this work introduces the preference alignment methods to the LM-based TTS systems. We demonstrate that the preference alignment methods boost the TTS system to outperform ground truth human speech in terms of speaker similarity, and proxy subjective evaluation scores. Exhaustive experiments are conducted to understand multiple critical issues in preference alignment implementation.


\bibliographystyle{IEEEbib}
\bibliography{ref}
\end{document}